\documentclass[sigconf]{acmart}

\setcopyright{none}
\renewcommand\footnotetextcopyrightpermission[1]{}
\makeatletter
\def\@copyrightspace{\relax}
\makeatother

\acmConference{}{}{}
\acmBooktitle{}
\acmYear{}
\copyrightyear{}
\acmISBN{}
\acmDOI{}

\AtBeginDocument{%
  }

\usepackage{multirow}
\usepackage[table,xcdraw]{xcolor}

\begin{document}

\title{Enabling Personalized Long-term Interactions in LLM-based Agents through Persistent Memory and User Profiles}

\author{Rebecca Westhäußer}
\email{rebecca.westhaeusser@mercedes-benz.com}
\orcid{0009-0008-1282-0811}
\affiliation{
    \institution{Mercedes-Benz AG}
    \city{Böblingen}
    \country{Germany}
}

\author{Wolfgang Minker}
\email{wolfgang.minker@uni-ulm.de}
\orcid{0000-0003-4531-0662}
\affiliation{
    \institution{Ulm University}
    \city{Ulm}
    \country{Germany}
}

\author{Sebastian Zepf}
\email{sebastian.zepf@mercedes-benz.com}
\orcid{0000-0002-1268-146X}
\affiliation{
    \institution{Mercedes-Benz AG}
    \city{Böblingen}
    \country{Germany}
}


\begin{abstract}
Large language models (LLMs) increasingly serve as the central control unit of AI agents, yet current approaches remain limited in their ability to deliver personalized interactions. While Retrieval Augmented Generation enhances LLM capabilities by improving context-awareness, it lacks mechanisms to combine contextual information with user-specific data. Although personalization has been studied in fields such as human-computer interaction or cognitive science, existing perspectives largely remain conceptual, with limited focus on technical implementation. To address these gaps, we build on a unified definition of personalization as a conceptual foundation to derive technical requirements for adaptive, user-centered LLM-based agents. Combined with established agentic AI patterns such as multi-agent collaboration or multi-source retrieval, we present a framework that integrates persistent memory, dynamic coordination, self-validation, and evolving user profiles to enable personalized long-term interactions.
We evaluate our approach on three public datasets using metrics such as retrieval accuracy, response correctness, or BertScore. We complement these results with a five-day pilot user study providing initial insights into user feedback on perceived personalization. The study provides early indications that guide future work and highlights the potential of integrating persistent memory and user profiles to improve the adaptivity and perceived personalization of LLM-based agents. 
\end{abstract}

\keywords{AI-Agents, Personalization, LLM}

\maketitle
\makeatletter
\def\@headfoot{}%
\def\@oddhead{}%
\def\@evenhead{}%
\makeatother

\section{Introduction}
Large Language Models (LLMs) demonstrate remarkable reasoning and planning capabilities that make them suitable as central control units for AI agents \cite{wu2024retrieval, singh2025agentic, acharya2025agentic, guo2024large}. These agents can autonomously perceive their surroundings, make decisions, and take actions in response \cite{guo2024large, newsham2025personality}. 

Despite this potential, current AI agents have several limitations, including static, task-specific designs, hallucination \cite{portugal2024agentic, sapkota2505ai, liu2023think}, and their lack of personalization \cite{yuan2023personalized, shan2025cognitive}. Furthermore, users often experience frustration when interacting with AI systems, as misunderstood queries lead to irrelevant responses \cite{shumanov2021making}. This stems from the inability of LLMs to retain and use past interaction data to deliver personalized responses \cite{pan2025memory}. This highlights the need for the ability to store and utilize information across interactions as a critical requirement for personalized interactions \cite{he2024human}.

Retrieval Augmented Generation (RAG) has become one of the primary methods for enhancing the capabilities of LLMs by retrieving relevant external information, but it does not address personalization \cite{zhou2024metacognitive, guo2024large, li2025survey, lewis2020retrieval, seo2025prompt}. Achieving personalization requires a user-specific, dynamically evolving memory and adaptation mechanism that aligns system behavior with human expectations \cite{hou2503model, agrawal2025scmrag, wang2023aligning, erdogan2024computational}. Recent studies highlight personalization as the next critical step for improving intelligent agents, as it enables more adaptive and context-aware AI systems \cite{li2025survey, wang2023aligning}, leading to more user acceptance, engagement, and interaction quality \cite{banna2025beyond, osman2024computational}.

While personalization has been studied in domains such as human-computer interaction and cognitive science, existing definitions mainly focus on user experience and behavioral outcomes \cite{fan2006personalization}. Less attention has been paid to a technical perspectives that link personalization to concrete system requirements for AI systems.

To address these gaps, we build on a unified definition of personalization as a conceptual foundation to derive technical requirements for adaptive, user-centered AI agents. We integrate these system requirements through six established agentic AI patterns such as Multi-Agent Collaboration or Multi-Source Retrieval, to present a framework that combines persistent memory, dynamic coordination, self-validation, and evolving user profiles to enable personalized interactions.
The key contributions of our approach are as follows:
\begin{itemize}
    \item We introduce a framework that applies derived technical requirements from our definition of personalization by combining established agentic AI patterns with evolving, user-specific memory modules and dynamic user profiling to enable adaptive, context-aware, and tailored interactions.
    \item We evaluate our framework on three public datasets and complement this with a five-day pilot user study, providing initial insights into users’ perception of personalization and adaptivity. The study provides early indications that guide future work and shows the potential of integrating persistent memory and user profiles to improve personalization.
\end{itemize}

This work is structured as follows: First, we review relevant existing work in the fields of personalization, memory-based AI systems, and established agentic AI patterns. Second, we define personalization, derive technical requirements, and introduce our system design. Third, we describe our experiments and present the resulting insights. Finally, we discuss our findings, draw conclusions, and outline potential directions for future work.

\section{Related Work}
\subsection{Personalization across Domains} 
Personalization has emerged as an essential capability in modern AI systems, enabling customized interactions that align with individual preferences, contexts, and goals \cite{li2025survey, osman2024computational}. Despite its growing importance, there is no agreed on definition on what defines personalization in AI systems from a technical perspective \cite{tiihonen2017introduction}, which complicates the design and evaluation of personalized AI systems.

Personalization has been interpreted differently across research domains. In early works it is defined as a process of changing a system to increase its personal relevance \cite{blom2000personalization}. From a cognitive science perspective, personalization refers to providing relevant content based on individual user preferences or behavior \cite{karat2000afford}. In human-computer interaction, personalization is defined as a way to close the gap between user and computer \cite{fan2006personalization}. In computer science, personalization refers to a toolbox of technologies to enhance user experience through system design \cite{kramer2000user}. More recently, in the context of AI, personalization is seen as the adaptation of LLMs to align with preferences, needs, and characteristics of an individual user or group of users \cite{zhang2024personalization, peng2025navigating}.

While these perspectives highlight the importance of personalization, they largely remain conceptual. They describe what personalization aims to achieve, but not how it should be realized technically. In particular, prior definitions rarely specify system-level features or mechanisms required to support personalization.
Addressing this gap, our work synthesizes existing definitions into a unified schema that connects conceptual perspectives with technical requirements, providing the foundation for our framework.

\subsection{Memory-based Personalized AI Systems}
RAG has emerged as a central approach for enhancing the capabilities of LLMs by integrating external resources into the generation process to produce more accurate and contextually aware outcomes \cite{zhou2024metacognitive, guo2024large, li2025survey, lewis2020retrieval, zhang2024personalization, wu2024retrieval, agrawal2025scmrag}.

Memory-augmented methods extend the RAG concept by incorporating contextual, user-specific information from past interactions \cite{seo2025prompt, westhausser2025caim}. One example is Prompt Chaining \cite{seo2025prompt}, which enhances long-term recall through multi-step reasoning combined with semantic similarity search. Similarly, the Cognitive AI Memory (CAIM) framework \cite{westhausser2025caim} introduces a tag-based retrieval process inspired by cognitive AI principles. More recently, Agrawal et al. \cite{agrawal2025scmrag} proposed SCMRAG, a self-corrective RAG system that integrates a dynamic LLM-based workflow.
Collectively, these approaches demonstrate foundational work and highlight the importance of combining retrieval mechanisms with memory modules to improve context-aware and personalized responses by leveraging information from prior interactions \cite{lewis2020retrieval, shan2025cognitive}.

These works focus on memory for information recall to improve context-awareness, but they struggle to adapt to evolving user contexts and remain limited in enabling personalized interactions \cite{hou2503model, agrawal2025scmrag}. Constrained by predefined reasoning steps and often relying on fragmented retrieval, they struggle with error detection and frequently generate responses that lack completeness \cite{zhou2024metacognitive, pan2025memory, westhausser2025caim}.

In contrast, our approach conceptualizes personalization not only as memory-augmentation. We combine persistent memory with evolving user profiles, dynamic coordination, and self-validation, thereby extending the retrieval-based context-awareness with adaptive, user-centered long-term interactions. This positions memory as one of several interconnected requirements for building personalized LLM-based agents.

\subsection{Agentic AI Patterns}
LLMs serve as the central control unit for AI agents \cite{wu2024retrieval, singh2025agentic, acharya2025agentic}. These agents are guided by prompts to solve specific tasks \cite{guo2024large, portugal2024agentic, sapkota2505ai}. More recently, research has shifted towards agentic AI, which refers to autonomous systems that can perceive, reason, act, and continuously learn from their environment, allowing them to dynamically optimize their performance \cite{zhang2025toward}. By collaborating with other agents and integrating real-time and historical context, agentic systems achieve high adaptability and context-aware decision-making \cite{portugal2024agentic, zhang2025toward}. This marks a key turning point from static, task-specific agents towards more dynamic agentic workflows \cite{sapkota2505ai}.

Recent work in the field of agentic AI has proposed a set of patterns that describe how AI agents can collaborate, reason, and adapt to evolving contexts \cite{singh2025agentic, sapkota2505ai}. These patterns provide design structures for agent-based systems and have been discussed across different domains \cite{acharya2025agentic, zhang2025toward, portugal2024agentic}. Six patterns have emerged as particularly relevant: (1) \textbf{Central Coordination}, to dynamically tailor retrieval strategies based on the complexity of the user query; (2) \textbf{Planning}, where agents decompose and organize tasks; (3) \textbf{Multi-Source Retrieval}, enabling the integration of external knowledge; (4) \textbf{Multi-Agent Collaboration}, where specialized agents collaborate to coordinate retrieval and reasoning processes; (5) \textbf{Reflection}, enabling iterative refinement of retrieval strategies; and (6) \textbf{Persistent Memory}, which maintains information across interactions.

While these patterns provide a conceptual foundation, their potential for enabling personalization has not been addressed. To our knowledge, prior work has primarily focused on agentic workflows for task efficiency and robustness \cite{zhang2025toward, singh2025agentic}, rather than adapting system behavior to individual users. We bridge this gap by connecting agentic patterns with the requirements of personalization, demonstrating how they support adaptive, user-centered systems.

\section{System Design}
\subsection{Defining Personalization}
Prior work has proposed multiple definitions of personalization, emphasizing different aspects such as content or behavior. To structure our approach we follow Fan et al. \cite{fan2006personalization} and synthesized existing definitions into an underlying schema that can be categorized into four dimensions: First, the \textit{object of personalization} refers to what aspect is personalized, such as content or functionality. Second, the \textit{target of personalization} states whether the adaptation is designed for an individual user or a group. Third, the \textit{actor of personalization} addresses whether personalization occurs implicitly through the system or explicitly through user input. Finally, the \textit{goal of personalization} highlights the intended outcome.

Building on this schema, we define personalization as the process of adapting a system’s output based on an understanding of the user's task, preferences, context, and historical interactions, in order to meet individual needs and enhance the user experience.

Following this definition, our approach aims for an individuated personalization, focusing on a specific user. Personalization in our system is both explicit - as the user contributes the content for personalization - and implicit - where the system autonomously decides what information to retain for future interactions. Overall, the system's personalization goal is to meet individual user needs to enhance the user experience.

Based on our definition, we derived concrete system requirements to enable personalization. First, achieving \textit{adaptivity} is enabled by an agentic workflow, in which specialized AI agents collaborate to understand the user's task and meet individual needs. Second, ensuring \textit{consistency} of conversational context across interactions is essential. We address this through user-specific memory modules based on RAG, which store and retrieve historical information. Third, delivering \textit{tailored responses} requires a user profile that captures key facts, including preferences, conversational patterns, and individual goals.
Together, the agentic workflow, memory modules, and user profiles provide the technical requirements for enabling personalization in our framework.

\subsection{Agentic Architecture}
\begin{figure*}[h!]
    \centering
    \includegraphics[width=1\linewidth]{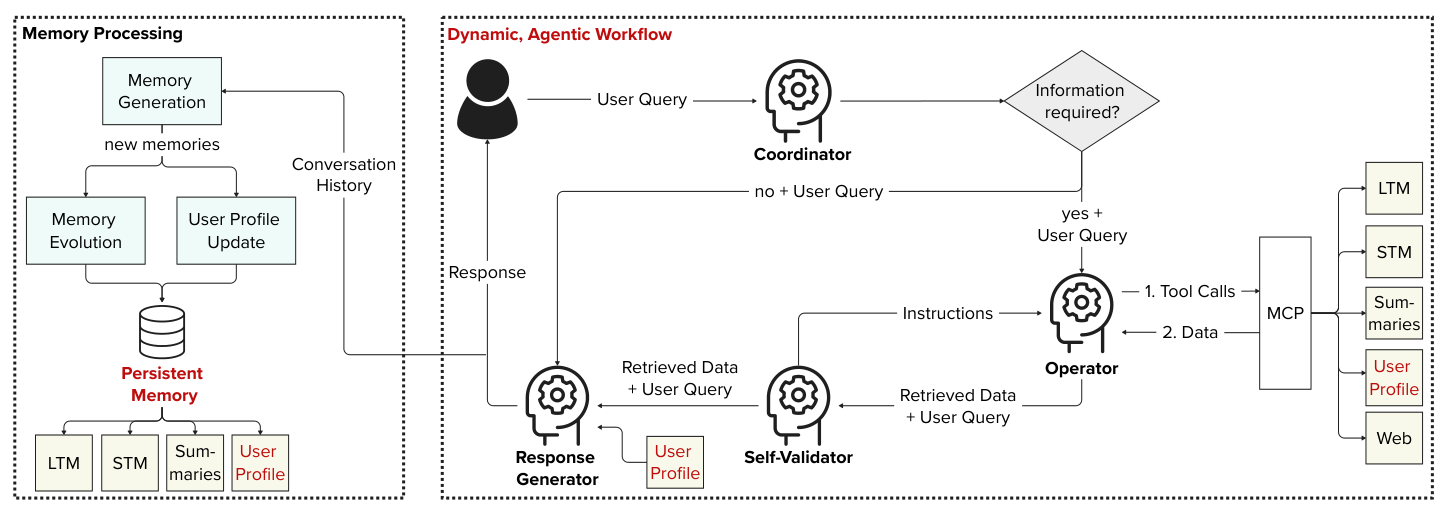}
    \caption{Agentic Workflow combining Agentic AI Patterns, Persistent Memory, and dynamic User Profiles}
    \Description{A diagram showing the workflow that combines Agentic AI Patterns, Persistent Memory, and dynamic User Profiles.}
    \label{fig:workflow}
\end{figure*}

Building on our definition of personalization, the first system requirement —adaptivity— requires mechanisms capable of flexibly responding to evolving user contexts. Agentic AI provides a structured framework for designing such mechanisms. In particular, we build upon six key patterns, namely Central Coordination, Planning, Multi-Source Retrieval, Multi-Agent Collaboration, Reflection, and Persistent Memory. While prior work has primarily linked them to task efficiency and robustness, we reinterpret them through the lens of personalization to provide a foundation for adaptive, user-centered system behavior. This reinterpretation is directly reflected in the system design, for example through role-specific prompts that reference user-specific information and interaction history.

As shown in Figure \ref{fig:workflow}, the agentic workflow starts with a user query, which is handled by the \textbf{Coordinator} agent. Following the Central Coordination pattern, the agent dynamically adapts retrieval strategies according to the complexity of the query \cite{singh2025agentic, acharya2025agentic}. For straightforward fact-based requests, the query is directly forwarded to the Response Generator. More complex queries that require additional context are forwarded to the Operator to get further information \cite{singh2025agentic}.
This pattern is essential for personalization, as the Coordinator tailors the system’s workflow to the individual user query, ensuring that retrieval and reasoning strategies are adapted to task complexity and user-specific context.

The second agent in our framework, the \textbf{Operator}, embodies the Planning pattern by interpreting the user query and autonomously selecting tools to process it. This enables dynamic, task-dependent retrieval strategies instead of relying on predefined, static steps \cite{portugal2024agentic, chen2017big, hou2503model, ferrag2025llm}.
To realize the Multi-Source Retrieval pattern, the Operator combines RAG with the Model Context Protocol (MCP), which is a standardized interface introduced by Anthropic \cite{Anthropic2025}. MCP provides access to external resources \cite{sapkota2505ai, singh2025agentic, hou2503model}, enabling the Operator to flexibly gather information beyond the model's pretrained knowledge. Through MCP, the Operator can use:
\begin{enumerate}
  \item a user-specific long-term memory (LTM) that stores interaction data as embeddings, which are retrieved via similarity search \cite{seo2025prompt, xu2025mem},
  \item a short-term memory (STM) for most recent conversational context \cite{westhausser2025caim},
  \item summaries capturing the main topics and key points from a wider span of past interactions, providing a broader view of the conversation history,
  \item a dynamic user profile containing key facts, and
  \item a web search API for retrieving the latest news \cite{ferrag2025llm, zhang2025toward}
\end{enumerate}

After retrieval, the Operator autonomously identifies the most relevant content regarding the user query and passes it to the following agent in the workflow \cite{zhang2025toward}.
This design is central for personalization: by flexibly deciding which tool to invoke, the Operator aligns retrieval with the user's evolving context, preferences, and interaction history. In contrast to static retrieval pipelines, this autonomous design supports adaptive, user-centered responses.

The following agent, the \textbf{Self-Validator}, embodies the Reflection pattern by integrating a self-critic mechanism to refine the models output \cite{agrawal2025scmrag, zhou2024metacognitive, sapkota2505ai}. Its primary task is to evaluate whether the information retrieved by the Operator is sufficient, consistent, and specific enough to answer the user query \cite{yuan2023personalized, acharya2025agentic, ferrag2025llm, beetz2007cognitive}. 
If gaps or inconsistencies are detected, the Self-Validator initiates a refinement process \cite{zhang2025toward, conway2024toward, toy2024metacognition}. The agent instructs the Operator on what additional information is required and how to improve retrieval. This iterative mechanism ensures that the retrieval is aligned with user-specific requirements, thereby overcoming the predefined strategies of existing RAG approaches.
The Self-Validator enforces a quality control layer that prevents incorporating insufficient information into the response \cite{banna2025beyond}. Furthermore, the agent ensures that the retrieved information is aligned with the current user context, supporting the generation of tailored responses.

Finally, the \textbf{Response Generator} transforms the retrieved information into a natural, user-friendly language \cite{portugal2024agentic}. The agent utilizes the user profile to adapt responses according to individual preferences and communication style. This step ensures that personalization is explicit at the interaction layer, producing outputs that are accurate, context-aware, and individually tailored.

Together these agents provide our system's foundation to adjust dynamically to evolving user-specific contexts \cite{sapkota2505ai, ferrag2025llm, zhang2025toward, li2025survey}. Within this workflow, the agentic patterns Central Coordination, Planning, Multi-Source Retrieval, Multi-Agent Collaboration, and Reflection address the derived system requirement of adaptivity.

\subsection{Memory Modules}
While adaptivity enables dynamic responses to evolving user contexts, personalization further requires the system to remember and reuse information across sessions. To realize the second system requirement —consistency— our framework integrates memory modules that maintain continuity over time. These modules provide the foundation for storing, organizing, and retrieving user-specific information from previous interactions, allowing the system to develop an evolving understanding of the user's preferences, goals, and communication patterns.

Therefore, the sixth agentic pattern we integrated is Persistent Memory, which stores and evolves information across interactions \cite{singh2025agentic, acharya2025agentic}. Following prior work, we use separated modules: a STM for recent conversational context, summaries capturing a broader view of the conversation history, a LTM for historical, user-specific data, and a user profile for easy access to key facts \cite{zhong2024memorybank, westhausser2025caim, he2024human, seo2025prompt, liu2023think}. This design allows the system to tailor responses based on both immediate context and historical, user-specific data, forming the foundation for adaptive, personalized interactions.

\subsubsection{\textbf{Short-term Memory}}
The STM stores the most recent conversational exchanges as full-text dialogue. Each message is initially stored within the STM before being summarized into broader topics. This module ensures that the system can generate contextually appropriate responses based on the immediate conversation flow.

\subsubsection{\textbf{Summaries}}
Summaries capture the main topics from a broader span of past interactions, providing an overview of conversation history beyond the STM. After a few exchanges, an LLM-based agent automatically generates these summaries to maintain awareness of the discussed topics.

\subsubsection{\textbf{Long-term Memory}}
The LTM stores user-specific, historical data.
To fill the LTM with newly generated memories, an LLM-based agent synthesizes the current conversation history into concise, context-aware summaries after every exchange, ensuring that only the most important information is retained \cite{zhang2025toward, yuan2023personalized}. Additionally, we implemented a tagging-based query expansion inspired by CAIM to semantically broaden the similarity search during retrieval with LLM-generated tags that capture the users intent or the main topic of the query \cite{westhausser2025caim, li2025survey, li2025knowledge}.
This combination of concise memories and tags ensures that only relevant information is stored and semantically linked, enabling a more nuanced understanding of relationships that goes beyond simple similarity metrics \cite{xu2025mem}.

Each stored memory consists of an embedding of the concise summary, LLM-generated tags, a timestamp, and a list of the top-5 related memories that are also retrieved to broaden the similarity search. To prevent outdated memories, the storage of a new memory automatically initiates an update of its top-5 related memories. This ensures that existing connections are continuously updated with new connections, mimicking human learning processes \cite{xu2025mem}.

Together, STM, summaries, and LTM form the core of personalization in our framework. By continuously integrating conversational and historical information, the system aligns its responses with the user's current context and interaction history. This evolving memory enables consistent, personalized interactions across sessions, thereby addressing the second system requirement derived from our definition of personalization.

\subsubsection{\textbf{User Profile}}
While the previous memory modules ensure continuity across interactions, personalization also requires a stable representation of the user to guide adaptation. To address the third system requirement —tailored responses— our framework integrates dynamic user profiles as part of the Persistent Memory, completing the technical foundation for personalization. These profiles contain key facts about the user and enable stable, user-centered personalization through tailored responses \cite{tang2025does}.

The user profile is implicitly generated and continuously refined by an LLM based on ongoing interactions. It includes key facts such as demographic information (e.g., name, age), user preferences, interests, and personality traits \cite{li2025survey, tang2025does}, as well as conversational characteristics such as tone or communication preferences. This approach is supported by prior work showing that everyday language use provides reliable insights into personality and communication preferences, which tend to remain consistent over time \cite{shumanov2021making}. 
At initialization, the user profile is represented as a structured JSON object containing predefined but empty categories. During interaction, an LLM-based agent receives the current conversation context and updates these categories with new information. This implicit and incremental profiling process enables continuous personalization by processing explicit dialogue context.

By integrating insights from past interactions into a stable representation, the user profile allows the system to generate responses that align with individual user traits and conversational tendencies. Prior work revealed that such adaptation enhances trust, engagement, and perceived personalization, emphasizing the central role of user modeling in human-computer interaction \cite{shan2025cognitive, shumanov2021making}.

Together, the agentic architecture, persistent memory, and user profile address the system requirements derived from our definition of personalization —adaptivity, consistency, and tailored responses— thus enabling a system that continuously learns from user interactions, maintains coherence over time, and adjusts its behavior to individual needs.

\section{Evaluation}
\subsection{Synthetic Evaluation}
\subsubsection{\textbf{Datasets}}
To evaluate our framework, we use three public datasets: Generated Virtual Dataset (GVD), LoCoMo, and LongMemEval. GVD contains conversation data involving 15 virtual users interacting with a virtual assistant over ten days, along with 100 English questions referencing conversation content \cite{zhong2024memorybank}. LoCoMo consists of ten long-term conversations, each between two users, and nearly 2000 questions, making it suitable for evaluating our systems ability to maintain consistency over extended dialogues \cite{xu2025mem, maharana2024evaluating}. LongMemEval evaluate five core LTM abilities of chat assistants: information extraction, multi-session and temporal reasoning, knowledge updates, and abstention \cite{wu2024longmemeval}. With these datasets, we evaluate our system's retrieval performance and how it affects response correctness and contextual coherence.

\subsubsection{\textbf{Metrics}}
Following prior work \cite{zhong2024memorybank, liu2023think, westhausser2025caim, zhang2019bertscore}, we adopt a set of metrics to evaluate the performance of our proposed system. The final set of metrics is as follows.
\begin{itemize}
    \item \textbf{Retrieval Accuracy} evaluates whether the relevant memories are successfully retrieved (labels: 0 = no; 1 = yes).
    \item \textbf{Response Correctness} evaluates whether the response contains the correct answer to the question (labels: 0 = wrong; 0.5 = partial; 1 = correct). 
    \item \textbf{Contextual Coherence} evaluates whether the response is naturally and coherently structured (labels: 0 = not coherent; 0.5 = partially coherent; 1 = coherent).
    \item \textbf{F1 - BertScore} computes token similarity.
    \item \textbf{ROUGE-1} measures lexical overlap via n-grams.
\end{itemize}

\subsubsection{\textbf{Baseline Model}}
To evaluate the effectiveness of our framework, we compare it against a standard RAG baseline and three ablation variants designed to isolate the contributions of key components —namely the Coordinator, the Self-Validator, and the User Profile.
The RAG baseline follows a conventional RAG setup, incorporating similarity-based retrieval over the entire dialogue history alongside the web search tool. In contrast our proposed framework operates on a structured, fragmented memory and additional processing agents. This design aims to enable long-term personalization and adaptive behavior, but it also introduces additional complexity that could affect performance.
Therefore, the goal of this evaluation is (1) to assess whether our framework maintains competitive retrieval and response quality compared to a RAG baseline, which is evaluated by using public datasets, and (2) to demonstrate the added value of our derived technical requirements in enabling adaptive, user-centered personalization, which is explored in the pilot user study to provide early indications to guide future work.

\subsubsection{\textbf{Evaluation Process}}
The evaluation is designed as a fact-checking task, determining whether our system retrieves the correct memories and provides accurate responses to the questions. For this purpose, we use GPT-4o as an evaluator. To validate our use of an LLM as the sole evaluator, we measured inter-rater agreement between three human annotators and GPT-4o on GVD, using it as a representative subset across the evaluation metrics. 
We calculated Percent Agreement (PA) to quantify how often the annotators agreed on the classification of each metric \cite{gisev2013interrater, chaturvedi2015evaluation}. The agreement between all raters was 93\% for retrieval accuracy, 81\% for response correctness, and 94\% for contextual coherence. In comparison, agreement among humans alone was 93\% for retrieval accuracy, 84\% for response correctness, and 97\% for contextual coherence. These results indicate that GPT-4o as an evaluator aligns with human judgments to a similar degree as humans align with each other. Based on these results, we consider GPT-4o as a reliable evaluator, which justifies its use for assessing the three datasets.

\subsubsection{\textbf{Results}}
We evaluate the performance of our system across the three synthetic datasets (GVD, LoCoMo, and LongMemEval) by comparing it to a RAG baseline and a series of ablation variants that remove the Coordinator, Self-Validator, and the User Profile. The results for each dataset and metric are summarized in Table \ref{tab:evaluation}.

\begin{table*}[h!]
    \caption{
        Experimental results on GVD, LoCoMo, and LongMemEval across five metrics using different models (GPT-4o, Gemini 2). The best performance is marked in bold, while grey cells highlight whether our approach or the baseline achieved higher performance.
    }
    \label{tab:evaluation}
    \begin{tabular}{ccl|ccccc}
        \hline
        \multicolumn{2}{c}{\textbf{Dataset}} & \textbf{Configuration} & \textbf{Retrieval Accuracy} & \textbf{Response Correctness} & \textbf{Contextual Coherence} & \textbf{BertScore} & \textbf{ROUGE-1} \\ \hline
        
         \multirow{10}{*}{\textbf{\rotatebox{90}{GVD}}} 
         & \multicolumn{1}{c|}{\multirow{5}{*}{\textbf{\rotatebox{90}{GPT-4o}}}} 
         & \textsc{RAG-Baseline} & 87\% & 81\% & \cellcolor[HTML]{d1d1e0}\textbf{98.5\%} & \cellcolor[HTML]{d1d1e0}\textbf{86.3\%} & \cellcolor[HTML]{d1d1e0}\textbf{19\%} \\
         & \multicolumn{1}{c|}{} & \textsc{Agentic System} & \cellcolor[HTML]{d1d1e0}\textbf{96\%} & \cellcolor[HTML]{d1d1e0}\textbf{90.5\%} & \cellcolor[HTML]{d1d1e0}\textbf{98,5\%} & 85.9\% & 17.4\% \\
         & \multicolumn{1}{c|}{} & \textsc{w/o Coordinator} & 92\% & 83.5\% & \textbf{98.5\%} & 85.7\% & 17.4\% \\
         & \multicolumn{1}{c|}{} & \textsc{w/o Self-Validator} & 95\% & 90\% & 96.5\% & 85.8\% & 16.9\% \\
         & \multicolumn{1}{c|}{} & \textsc{w/o User Profile} & 91\% & 88\% & 97\% & 85.8\% & 17.5\% \\ \cline{2-8}
         
         & \multicolumn{1}{c|}{\multirow{5}{*}{\textbf{\rotatebox{90}{Gemini 2}}}} 
         & \textsc{RAG-Baseline} & \cellcolor[HTML]{d1d1e0}87\% & \cellcolor[HTML]{d1d1e0}76\% & 96.5\% & \cellcolor[HTML]{d1d1e0}\textbf{86,6\%} & \cellcolor[HTML]{d1d1e0}\textbf{23,9\%} \\
         & \multicolumn{1}{c|}{} & \textsc{Agentic System} & 85\% & \cellcolor[HTML]{d1d1e0}76\% & \cellcolor[HTML]{d1d1e0}97\% & 85.9\% & 20.4\% \\
         & \multicolumn{1}{c|}{} & \textsc{w/o Coordinator} & \textbf{91\%} & \textbf{81.5\%} & 97\% & 85.2\% & 20.2\% \\
         & \multicolumn{1}{c|}{} & \textsc{w/o Self-Validator} & 86\% & 79.5\% & \textbf{99\%} & 85.8\% & 19.8\% \\
         & \multicolumn{1}{c|}{} & \textsc{w/o User Profile} & 81\% & 74\% & 96.5\% & 85.9\% & 20.9\% \\ \hline
        
         \multirow{10}{*}{\textbf{\rotatebox{90}{LoCoMo}}} 
         & \multicolumn{1}{c|}{\multirow{5}{*}{\textbf{\rotatebox{90}{GPT-4o}}}} 
         & \textsc{RAG-Baseline} & 62.1\% & 58.9\% & \cellcolor[HTML]{d1d1e0}\textbf{98.5\%} & 82.5\% & 6.5\% \\
         & \multicolumn{1}{c|}{} & \textsc{Agentic System} & \cellcolor[HTML]{d1d1e0}83.4\% & \cellcolor[HTML]{d1d1e0}78.8\% & 93.5\% & \cellcolor[HTML]{d1d1e0}82.7\% & \cellcolor[HTML]{d1d1e0}7.3\% \\
         & \multicolumn{1}{c|}{} & \textsc{w/o Coordinator} & 83.6\% & 75.3\% & 96.1\% & 82.7\% & 7.2\% \\
         & \multicolumn{1}{c|}{} & \textsc{w/o Self-Validator} & \textbf{88.7\%} & \textbf{81\%} & 93.2\% & \textbf{82.9\%} & 7.6\% \\
         & \multicolumn{1}{c|}{} & \textsc{w/o User Profile} & 81.7\% & 77.4\% & 98.4\% & \textbf{82.9\%} & \textbf{ 8.1\%} \\ \cline{2-8}
         
         & \multicolumn{1}{c|}{\multirow{5}{*}{\textbf{\rotatebox{90}{Gemini 2}}}} 
         & \textsc{RAG-Baseline} & 59.7\% & 55.6\% & 91.6\% & \cellcolor[HTML]{d1d1e0}\textbf{83.1\%} & \cellcolor[HTML]{d1d1e0}\textbf{9.7\%} \\
         & \multicolumn{1}{c|}{} & \textsc{Agentic System} & \cellcolor[HTML]{d1d1e0}69.1\% & \cellcolor[HTML]{d1d1e0}59.5\% & \cellcolor[HTML]{d1d1e0}\textbf{93\%} & 82.9\% & 8.4\% \\
         & \multicolumn{1}{c|}{} & \textsc{w/o Coordinator} & \textbf{80.3\%} & \textbf{72.6\%} & 92.7\% & \textbf{83.1\%} & 9.6\% \\
         & \multicolumn{1}{c|}{} & \textsc{w/o Self-Validator} & 65.8\% & 57\% & 92.6\% & 82.8\% & 8\% \\
         & \multicolumn{1}{c|}{} & \textsc{w/o User Profile} & 65.1\% & 54.9\% & 91.7\% & 82.9\% & 8.8\% \\ \hline
        
         \multirow{10}{*}{\textbf{\rotatebox{90}{LongMemEval}}} 
         & \multicolumn{1}{c|}{\multirow{5}{*}{\textbf{\rotatebox{90}{GPT-4o}}}} 
         & \textsc{RAG-Baseline} & 84.2\% & \cellcolor[HTML]{d1d1e0}\textbf{77.7\%} & 97.4\% & \cellcolor[HTML]{d1d1e0}\textbf{82.7\%} & \cellcolor[HTML]{d1d1e0}\textbf{9.2\%} \\
         & \multicolumn{1}{c|}{} & \textsc{Agentic System} & \cellcolor[HTML]{d1d1e0}\textbf{87.2\%} & 72.8\% & \cellcolor[HTML]{d1d1e0}\textbf{98.3\%} & 82.4\% & 8.2\% \\
         & \multicolumn{1}{c|}{} & \textsc{w/o Coordinator} & 84\% & 69.7\% & 98.1\% & 82.2\% & 8\% \\
         & \multicolumn{1}{c|}{} & \textsc{w/o Self-Validator} & 83.4\% & 68.3\% & 96.2\% & 82.3\% & 8\% \\
         & \multicolumn{1}{c|}{} & \textsc{w/o User Profile} & 81.6\% & 70\% & 98.1\% & 82.2\% & 7.8\% \\ \cline{2-8}
         
         & \multicolumn{1}{c|}{\multirow{5}{*}{\textbf{\rotatebox{90}{Gemini 2}}}} 
         & \textsc{RAG-Baseline} & 82\% & 65.7\% & 91.6\% & \cellcolor[HTML]{d1d1e0}\textbf{83.2\%} & \cellcolor[HTML]{d1d1e0}\textbf{11.4\%} \\
         & \multicolumn{1}{c|}{} & \textsc{Agentic System} & \cellcolor[HTML]{d1d1e0}\textbf{82.6\%} & \cellcolor[HTML]{d1d1e0}68.2\% & \cellcolor[HTML]{d1d1e0}92\% & 83\% & 10.8\% \\
         & \multicolumn{1}{c|}{} & \textsc{w/o Coordinator} & 82.2\% & \textbf{68.8\%} & 91.6\% & 83.1\% & 11\% \\
         & \multicolumn{1}{c|}{} & \textsc{w/o Self-Validator} & 76\% & 67.2\% & \textbf{94.2\%} & 82.9\% & 9.9\% \\
         & \multicolumn{1}{c|}{} & \textsc{w/o User Profile} & 75.4\% & 65.5\% & 89.5\% & 82.9\% & 10\% \\ \hline
    \end{tabular}%
\end{table*}

\paragraph{\textbf{Overall Performance}}
Across datasets, our framework improves retrieval accuracy and response correctness compared to the RAG baseline, while maintaining comparable contextual coherence. The results confirm that integrating agentic patterns such as coordination and validation, and a user modeling mechanism enables more reliable retrieval and response correctness, even in synthetic, non-adaptive settings.

\paragraph{\textbf{Comparison to RAG Baseline}}
Across datasets, our agentic framework achieves competitive or superior performance compared to the RAG baseline. For GPT-4o, retrieval accuracy improved by up to 20\% on LoCoMo, 11\% on GVD, and 3\% on LongMemEval, with similar results in response correctness.
For Gemini 2, the results follow a similar pattern. Our approach outperforms the baseline by 10\% in retrieval accuracy and 4\% in response correctness on LoCoMo, while performance remains comparable on GVD and LongMemEval.
These findings confirm that our approach generalizes across model architectures and maintains strong retrieval and response capabilities, particularly in tasks requiring contextual adaptation and user-centered reasoning.

\paragraph{\textbf{Ablation Study}}
Removing the Coordinator leads to mixed effects across models and datasets. For GPT-4o, the absence of this component consistently reduced both retrieval accuracy (up to -4\%) and response correctness (up to -7\%). This indicates that the Coordinator effectively manages the retrieval of additional information, ensuring  that relevant context is integrated into the response generation process. Even when retrieval accuracy remains stable (e.g. LoCoMo), response quality decreases, suggesting that coordination also benefits contextual reasoning and coherence rather than retrieval alone.
In contrast the Gemini 2 model shows opposite trends. On GVD and LoCoMo, removing the agent improved performance between 5.5\% and 13\%. This effect stems from model-specific behavior. When the Coordinator was included, Gemini 2 often refused to use the provided memory modules despite explicit instructions, leading to incomplete context integration. Removing the agent bypassed this limitation, resulting in more consistent and contextually grounded responses.
Overall, these results highlight that the benefit of the Coordinator depends on the underlying models abilities.

Excluding the Self-Validator generally leads to minor changes in performance across models and datasets. For GPT-4o, the removal had little to no impact, suggesting that the agent occasionally introduces complexity by rejecting valid retrieval results. However, on LongMemEval, both retrieval accuracy and response correctness decreased by 4\%, demonstrating that the Self-Validator contributes to maintaining consistency when processing fragmented memory contexts.
For Gemini 2, the effects follow a similar pattern. The performance remains largely stable or slightly decreases (between 2\% and 6\%), indicating that the Self-Validator contributes only slightly to overall system performance under controlled evaluation settings. However, its role could be more relevant for noisy, human-generated data.

Removing the User Profile reduced retrieval accuracy and response correctness across all datasets and models. For GPT-4o, retrieval accuracy decreased by up to 5.6\% and response correctness by 2.8\%, indicating that even while using synthetic data the profile contributes to retrieval and generating more relevant, user-aligned responses.
A similar trend was observed for Gemini 2, with retrieval accuracy decreasing between 4\% and 7\% and response correctness by up to 4.6\%.
These results highlight the importance of an explicit user representation for ensuring stable user modeling and long-term personalization within our agentic framework.
Therefore, the findings confirm that the User Profile contributes not only to coherent user modeling but also to overall system performance, effectively realizing our system requirement of tailored responses.

\paragraph{\textbf{BertScore and ROUGE-1}}
BertScore and ROUGE-1 show limited alignment with human-level evaluation objectives. ROUGE-1 fails to capture semantically correct responses that differ from reference outputs. For example, our agentic system achieves lower ROUGE-1 scores than the baseline while providing more accurate and contextually coherent answers.
Similarly, BertScore, though sensitive to token-level semantic similarity, does not reliably reflect correctness or contextual coherence in dialogue, as the metric consistently ranges between 82\% and 86\% across configurations. These observations support our decision to prioritize human-level judgment using metrics such as retrieval accuracy, response correctness, and contextual coherence.

The evaluation on synthetic datasets confirms that our agentic framework maintains competitive performance compared to the RAG baseline, particularly in retrieval accuracy and response correctness. The ablation studies indicate that the User Profile is essential for preserving personalized context, whereas the Coordinator and Self-Validator contribute in more nuanced, dataset-dependent ways. These results confirm that our agentic framework remains competitive in retrieval, response generation, and coherence, providing the foundation for adaptivity as defined in our system requirements.
However, since synthetic datasets cannot fully represent human preference diversity, we extend our evaluation with a pilot user study to assess real-world personalization and adaptivity.

\subsection{Pilot User Study}
While the evaluation on synthetic datasets enables controlled testing of individual system components, it cannot capture personalization, nuanced user preferences, or dynamic conversational behavior. As this paper primarily focuses on the technical realization and evaluation of personalization mechanisms, we conducted a five-day pilot study to complement our findings with initial insights into users’ perception of personalization and adaptivity. The study provides early indications that will guide future work, in which we plan to conduct a longitudinal user study to systematically evaluate personalization effects.

\subsubsection{\textbf{Setup}}
We recruited seven participants with diverse professional backgrounds. Each participant installed our AI chatbot as a mobile application on their personal smartphone and interacted with it over five consecutive days. Each participant received an onboarding session explaining the chatbot’s general functionality without revealing technical details about its architecture to minimize bias. Participants were instructed to choose a topic of personal interest to converse with the system for around ten minutes per day. After each daily session, participants answered questionnaires about their experience and perceived system behavior.

The mobile application is designed as an AI-chatbot based on our agentic framework. For each user query, the Coordinator decides whether additional information is required. If necessary, the Operator retrieves relevant contextual data. The Self-Validator evaluates whether the retrieved information is sufficient and consistent. Finally, the Response Generator produces a response based on the retrieved content and informed by the user profile. 
We chose a mobile application as the evaluation environment because it supports natural, everyday interactions over multiple days.

\subsubsection{\textbf{Study Results}}
Prior work has stated that there is no agreed on method to evaluate personalization in systems \cite{tiihonen2017introduction}. Since personalization is subjective and context-dependent, we focused on user's perception to assess whether our system's personalization mechanisms were recognized and appreciated. Therefore, participants were asked to reflect on their experience and to evaluate the chatbot's perceived personalization. 

Out of seven participants, four completed the five-day study, while three discontinued after the first day. The early dropout highlights the challenge of maintaining engagement over extended periods and indicates the need for motivational elements in future longitudinal studies. The remaining participants provided valuable insights into perceived personalization of our agentic framework.

Overall, participants described the interaction as natural and increasingly personalized over time. Two participants described the chatbot's language as "human-like", noting that responses followed a natural conversational flow. All four participants highlighted the system's ability to recall previous conversations as a key element of personalization. References to past interactions and the reuse of user-stated preferences supported the impression of being remembered by the system. This insight confirms that the persistent memory component effectively contributes to long-term consistency, thereby realizing the second system requirement of our personalization framework.
One participant further noted that the contextual awareness improved throughout the study, as the system increasingly integrated previous conversation topics and refined its focus around the user's chosen topic of interest. This indicates that the persistent memory and user profile components contributed to an evolving, user-centered experience.

Despite these positive aspects, participants also noted several limitations. Three users criticized the chatbot’s responses as long and generic. Similarly, three participants reported issues with ambiguous queries such as vague follow-up questions ("Is it helpful?") where the system failed to utilize prior conversational context. This indicates that the contextual reasoning processes require refinement to better handle vague user input.

Participants also provided constructive suggestions for improving personalization. Two participants emphasized the need for proactive behavior, suggesting that the chatbot should take initiative and guide the conversation, for example by asking follow-up questions. One participant suggested adding customization options to adjust the chatbot's conversational style and response behavior. Another participant proposed an onboarding phase to explicitly collect personal background information, enabling the system to develop a more holistic and individualized interactions style.

Overall, the pilot study provides initial insights about users recognizing and appreciating the system’s personalization mechanisms. At the same time, the results reveal the need for improvements in contextual reasoning, more engaging dialogue strategies, and proactive interaction design. Most notably, the high initial dropout rate emphasizes that long-term studies on personalization require mechanisms that maintain user motivation over time.

\section{Discussion}
Our results reveal key insights into the design and challenges of personalized conversational AI systems, aligning closely with findings in recent literature. By combining our evaluation results, we are able to analyze the extent to which the proposed system requirements —adaptivity, consistency, and tailored responses— were realized in practice and how they could be improved.

Adaptivity was partially achieved through the system's ability to adjust its responses to evolving conversational contexts. In the synthetic analysis, the ablation results indicate that the Coordinator and Self-Validator improved performance in tasks requiring contextual reasoning, confirming their contribution to adaptive behavior. 
Participants in the pilot study similarly noted that the system increasingly referenced prior discussed information and refined its focus around the user's topic of interest. Therefore, our system adapts based on explicit contextual cues within the conversation.
While our proposed framework primarily reacts to user input, truly agentic systems initiate interactions based on contextual cues or learned patterns \cite{sapkota2505ai}. The participants indicated that such proactive behavior would enhance motivation and engagement, as the system would be perceived as more collaborative and present. In our context, proactive dialogue could improve the learning-oriented nature of the interaction by helping users to explore related topics beyond their initial focus.

Consistency emerged as one of the system's strengths in both evaluations. The synthetic evaluation mostly showed stable retrieval and response correctness across datasets, while user feedback highlighted that references to previous discussed topics created a sense of continuity and coherence. This perception stems from our integrated memory modules, which allow the system to reuse historical and user-related information over extended periods. 
Even though the system's responses contained relevant references, the participants found the chatbot's linguistic style too generic and noted that it is "LLM-like", which reduced the perceived naturalness of the conversation. This highlights the importance of adjusting the conversational style according to user preferences. Achieving both will be essential for human-aligned personalization.

The user profile contributed to tailored responses by maintaining a consistent representation of the user's preferences and goals. Participants noted that prior discussed topics influenced future responses, reinforcing the impression of being remembered by the system. The ablation studies confirmed this effect quantitatively, showing that integrating the user profile improves overall performance across all metrics, highlighting its central role for generating tailored responses.
One participant further suggested that explicitly incorporating personal background information during an onboarding phase could further improve the perceived personalization. Prior work similarly highlights that integrating user values and personality traits leads to higher acceptance, engagement, and interaction quality \cite{banna2025beyond, osman2024computational, erdogan2024computational}. Explicitly incorporating this information could therefore complement the system’s adaptive mechanisms, enable more tailored interactions, and contribute to more authentic, human-aligned personalization. 

Despite positive trends across all three system requirements, several challenges persist. A well-documented limitation of personalized AI systems is the cold-start problem, which arises when systems have limited historical data \cite{zhang2024personalization}. Our agentic framework cannot utilize user-specific data without prior interactions, leading to initially generic responses. This was also observed in the pilot study, where personalization effects became noticeable only after several interactions. Given that our study lasted five days, the short duration likely constrained the development of deeper personalization. Consequently, a follow-up longitudinal study is planned to investigate how using the AI-chatbot over several weeks affects more stable user modeling and personalized interactions.
Furthermore, the limited proactivity and generic response style indicate that providing relevant contextual responses alone does not guarantee perceived personalization or naturalness. In particular, initiating contextually meaningful dialogues and adjusting to the user's preferred conversational style could close the gap between functional personalization and human-like interactions. 

Our results indicate that personalization is not a static capability but an evolving process shaped by continuous interactions and adaptive memory mechanisms. At the same time, several limitations observed in our pilot study reflect well-known problems in the literature and point towards the need for more proactive, human-aligned system behavior.

\section{Conclusion}
In this work, we proposed an agentic framework for personalized conversational AI systems, based on a unified definition of personalization. From this definition, we derived three core system requirements — adaptivity, consistency, and tailored responses — and implemented them through an AI workflow, based on agentic design patterns, a persistent memory, and a dynamic user profile.

Our evaluation confirmed that each system component contributes to realizing one of the defined requirements: the agentic workflow supports adaptive behavior, the persistent memory enables consistency across interactions, and the user profile ensures tailored responses. On public datasets, the framework maintained competitive performance compared to RAG baselines while achieving higher stability through explicit user modeling. 
The pilot study revealed that users perceived and appreciated the system's personalization capabilities, particularly its ability to recall and integrate past interaction data. However, the study provides early indications that guide future work, revealing areas for improvement, such as the need for proactivity and improving contextual reasoning in real-world use cases. 
Overall, our findings show that personalization is an evolving process, requiring continuous interaction, dynamic memory integration, and alignment with individual user values.

\section{Future Work}
Building on the findings of this work, future research will focus on extending the proposed agentic framework towards more proactive, human-aligned personalization. A primary direction is to address the cold-start problem by enabling the system to collect user preferences, values, personality traits, and goals earlier through a lightweight onboarding phase.
Furthermore, we plan to strengthen the system’s proactive capabilities by enabling the agents to actively guide the conversation through reflective or exploratory dialogues.

Finally, future work will include a longitudinal user study that builds upon the findings from our pilot study to investigate how perceived personalization evolves over weeks of continuous use. This will provide a deeper understanding of how agentic systems can foster meaningful, human-centered relationships through adaptive and memory-driven behavior, ultimately aiming to meet individual needs and enhance the overall user experience. 

\bibliographystyle{ACM-Reference-Format}
\bibliography{sample-base}

\end{document}